\documentclass{article}

\usepackage[utf8]{inputenc}
\usepackage{natbib}
\usepackage{graphicx}
\usepackage[a4paper, total={6.5in, 8in}]{geometry}
\usepackage[position=b]{subcaption}
\usepackage{authblk}

\usepackage{amsmath, amsfonts, amsthm, amssymb}
\usepackage{mathtools}
\usepackage{xcolor}
\usepackage{enumitem}

\RequirePackage{fix-cm}

\DeclarePairedDelimiter\abs{\lvert}{\rvert}%
\DeclarePairedDelimiter\norm{\lVert}{\rVert}%

\makeatletter
\let\oldabs\abs
\def\abs{\@ifstar{\oldabs}{\oldabs*}}
\let\oldnorm\norm
\def\norm{\@ifstar{\oldnorm}{\oldnorm*}}
\makeatother

\theoremstyle{remark}
\theoremstyle{definition}
\usepackage{algorithm, algorithmic}

\def\Normal{{\mathcal{N}}}
\def\RR{{\mathbb{R}}}
\def\EE{{\mathbb{E}}}

\newtheorem{theorem}{Theorem}[section]
\newtheorem{lemma}[theorem]{Lemma}
\newtheorem{proposition}[theorem]{Proposition}
\newtheorem{definition}[theorem]{Definition}

\def\State{{\cal S}}

\def\Action{{\cal A}}
\def\Normal{{\cal N}}

\def\KL{\mathrm{KL}}

\def\R{\mathbb{R}}

\DeclarePairedDelimiterX{\KLD}[2]{(}{)}{%
  #1\;\delimsize\|\;#2%
}
\newcommand{\KLM}{\KL\KLD}

\begin{document}
	
\title{Convex Optimization with an Interpolation-based Projection and its Application to Deep Learning}


\author[1]{Riad Akrour}
\author[2]{Asma Atamna}
\author[1]{Jan Peters}
\date{}
\affil[1]{TU Darmstadt}
\affil[2]{Télécom Paris}

\maketitle

\begin{abstract}
Convex optimizers have known many applications as differentiable layers within deep neural architectures. One application of these convex layers is to project points into a convex set. However, both forward and backward passes of these convex layers are significantly more expensive to compute than those of a typical neural network. We investigate in this paper whether an inexact, but cheaper projection, can drive a descent algorithm to an optimum. Specifically, we propose an interpolation-based projection that is computationally cheap and easy to compute given a convex, domain defining, function. We then propose an optimization algorithm that follows the gradient of the composition of the objective and the projection and prove its convergence for linear objectives and arbitrary convex and Lipschitz domain defining inequality constraints. In addition to the theoretical contributions, we demonstrate empirically the practical interest of the interpolation projection when used in conjunction with neural networks in a reinforcement learning and a supervised learning setting.
\end{abstract}

\section{Introduction}
Several recent research has investigated the integration of a `convex optimization layer' within the computational graph of machine learning architectures in applications such as optimal control \citep{deavila2018, amos2018}, computer vision \citep{bertinetto2018, lee2019} or filtering \citep{barratt2019}. Within this line of research, we distinguish two use cases for convex optimization. In the first use case, the output of the `convex optimization layer' is a convex problem by definition. For example, a node can compute the maximum a posteriori of an image model \citep{deavila2018, amos2018}. In the second use case, a node restricts---by means of a projection---its input to a convex set and becomes a convex optimization problem by choice. For example, a node can restrict its input to the set of physically plausible vertex deformations \citep{geng2019}. 

In the second use case, it was shown in \cite{geng2019} that the projection step benefits from being fully integrated to the learning process in both the forward and backward passes. Letting $x$ be the input of the projection layer, $g$ be the projection, and $f$ be the ensuing computations---e.g. a loss function; integrating the projection into the backward pass amounts to differentiating through $f\circ g (x)$. There have been several advances in differentiating through convex programs \citep{agrawal2019}. However, the forward and backward passes on $g$ remain significantly more expensive than the typical matrix multiplications that would precede or succeed $g$ \citep{amos2017}. We investigate in this paper an alternative projection that is more lightweight to compute and differentiate than solving a convex program. Even if sub-optimal, in the sense that the proposed projection will not return the closest point to the input within the admissible set, the rationale behind the proposed algorithm is that since we are differentiating through $f$ \textit{and} $g$, a sub-optimal projection could still drive the optimization process to an optimal point.

The proposed projection maps any input $x$ to a feasible point $g(x)$ by simply interpolating $x$ with a point $x_0$ satisfying the convex inequality constraints. The interpolation parameter is computed in closed form by exploiting the convexity of the domain defining function. We first show in this paper that the interpolation-based projection when used as in Projected Gradient Descent \citep{rosen1960,nocedal2006}---by projecting the iterate after each gradient step---does not converge to an optimum. However, when differentiating through both the objective and the projection, we show that the resulting algorithm converges for a linear objective and arbitrary convex and Lipschitz domain defining functions. Finally, we provide in addition to the theoretical analysis, empirical results using the projection in conjunction with neural network models in reinforcement and supervised learning. Our results show that the proposed projection can be used to tackle constrained policy optimization or to provide an inductive bias improving generalization while being significantly cheaper to compute than an orthogonal, `optimal' projection.

This work generalizes and formally analyzes previous interpolation-based projections we developed in the context of reinforcement learning (RL) in \cite{Akrour19}. Several RL algorithms add information-theoretic constraints  to the policy optimization problem, such as a minimal entropy or a maximal Kullback-Leibler~(KL) divergence to the data generating policy \citep{Deisenroth13}. We proposed in \cite{Akrour19} differentiable policy parameterizations that comply with these constraints by construction, allowing the policy optimization problem to be solved by standard gradient descent algorithms. These parameterizations were based on interpolating any input parameterization of a distribution with a constraint satisfying parameterization. For example, interpolating an input discrete distribution with the uniform distribution, that satisfies any reasonable minimal entropy constraint. Interestingly, albeit these projections were not `optimal' in the sense that they do not minimize a distance to the admissible set, we noted empirically (see \cite{Akrour19}, Fig. 1 and surrounding text) that such parameterization would always drive the descent algorithm to an optimum on a toy problem with a linear objective and a convex, entropy constraint. The main contribution of this paper is to generalize the idea of interpolation projections to arbitrary convex domain defining functions and to prove convergence of a descent algorithm leveraging this projection. From a practical point of view, in addition to the previously discussed RL application, we provide an example usage of the interpolation projection in a supervised learning context. The interpolation projection can be used as an inexpensive and differentiable operator to add convex constraints to the output of a neural network model, while being significantly cheaper than norm minimizing projections~\citep{agrawal2019}.

Computationally frugal projections were previously studied in the context of feasibility problems \citep{Combettes97}, where the goal is to find a point inside a convex set. The approximate projection in \cite{Combettes97} uses the gradient of a violated inequality constraint to find a half-space that is a superset of the feasible set. Then an orthogonal projection on this hyper-plane is performed resulting in a point outside of the feasible set, but closer to the set than the input point. In contrast, our projection is not based on the gradient of the constraint but on its convexity and results in a point inside the feasible set. Moreover, the optimization setting we consider is more general than the feasibility setting and our assumption of an initial feasible $x_0$ would already solve the problem of~\cite{Combettes97}. As such, our work and that of \cite{Combettes97} differ both in their objectives and their methods.  In \cite{xu2018primaldual, lan2016algorithms}, approximate projections are derived when the number of constraints is large, but these algorithms still rely on expensive orthogonal projections. To the best of our knowledge, no other work previously showed convergence of a convex optimizer with non-orthogonal projections. The practical implications being a cheap way of adding convex constraints to machine learning models as shown in the experimental validation section.

\section{Preliminaries}
\label{sec:prelim}
Let us first introduce and analyse the ideas in a convex optimization setting. Let $f : \RR^d \rightarrow \RR$ and $h: \RR^d \rightarrow \RR$ be convex and differentiable functions. We consider the following convex program
\begin{equation}
\begin{aligned}
\min_{x \in \RR^d} \quad &	 f(x),\\
\text{s.t.} \quad & h(x) \leq 0.
\end{aligned}
\tag{P}\label{eq:pbm}
\end{equation}

For clarity of exposition, we initially only consider a single inequality constraint with differentiable $h$. Our results will be straightforwardly extended to multiple inequality constraints in Sec.~\ref{sec:sub} with sub-differentiable functions. Letting the convex set ${\cal C} \subseteq \RR^d$ be defined by ${\cal C}=\{x\in\RR^d: h(x) \leq 0\}$, the optimization problem \eqref{eq:pbm} can be reformulated as $\min_{x\in\cal C}f(x)$. To solve this problem, one approach is to use the Projected Gradient Descent (PGD) algorithm~\citep{rosen1960,nocedal2006} which is given by the following equation
\begin{align}
x_{k+1} &= g\left(x_k -\alpha\nabla f(x_k)\right),\label{eq:pgd}
\end{align}
where $g$ is a mapping that projects points from $\RR^d$ to $\cal C$. The projection $g$ is defined by the minimization $g(x) = {\arg\min}_{y\in {\cal C}} \norm{x-y}_2$ of the Euclidean norm $\norm{.}_2$ on $\RR^d$. Mirror descent \citep{bubeck2014convex}, an alternative for solving \eqref{eq:pbm}, can be seen as a generalization of PGD to other distances. These projection-based methods are most efficient when a closed form expression of the projection exists. Otherwise, a nested optimization problem needs to be solved after every gradient update of the iterate. 

Other approaches such as the Frank-Wolfe method or the interior-point method also solve series of optimization problems. The Frank-Wolfe method~\citep{frank1956} solves a series of linear approximations of the problem, $x_{k+1} = \arg\min_{x \in \cal C}\nabla f(x_k)^Tx$; and the interior-point method~\citep{karmarkar1984,nesterov1994}
introduces a slack variable $s$ for the inequality constraint and solves $f(x) - \mu_k \ln s$  under an equality constraint, for a series of values of $\mu_k$ going to 0. 

In contrast to all these methods, our algorithm takes a simpler and more direct approach by performing gradient descent on the composition of the objective and a projection. The proposed interpolation-based projection will transform the constrained problem \eqref{eq:pbm} into an unconstrained one. The projection is readily defined without any other assumption than the convexity of $h$. Unlike previous algorithms, the interpolation projection is not defined as the minimization of a norm. To alleviate any ambiguity, from here on the term projection is understood as the more general following definition.
\begin{definition}
A projection $g$ is a mapping from a set to a subset thereof.
\label{def:proj}
\end{definition}
Specifically, in this paper the superset is $\RR^d$ and the subset is $\cal C$. 
\section{Interpolation-based projection and gradient descent}
\label{sec:alg}
\begin{figure}
	\center
	\begin{subfigure}[b]{0.3\textwidth}
		\includegraphics[width=\textwidth]{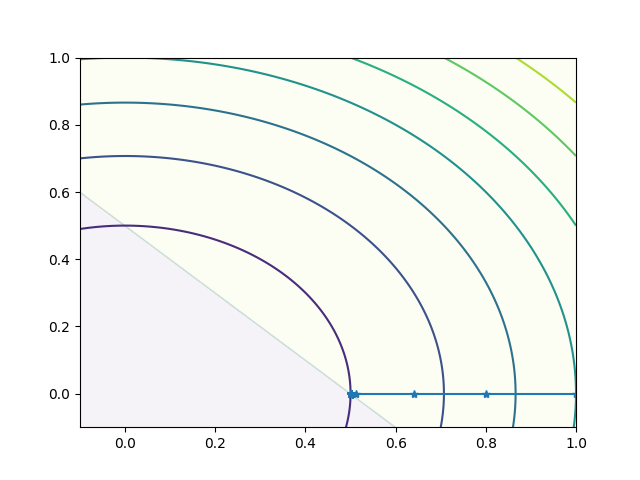}
		\caption{$x_{k+1} = g(x_k - \alpha \nabla f(x_k))$}
	\end{subfigure}
	\begin{subfigure}[b]{0.3\textwidth}
		\includegraphics[width=\textwidth]{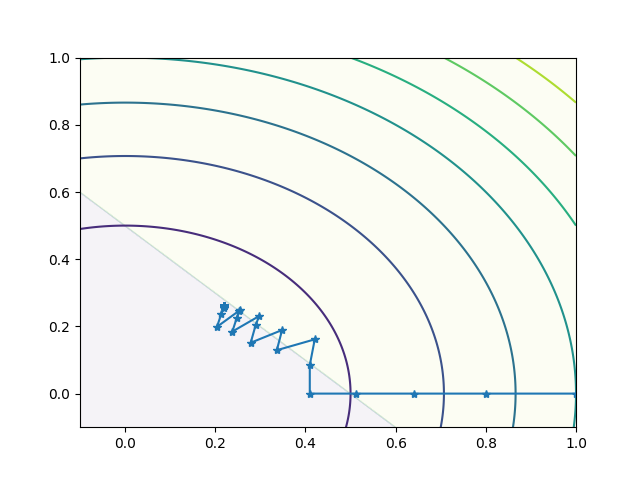}
		\caption{$x_{k+1} = x_k - \alpha \nabla f\circ g (x_k)$}
	\end{subfigure}
	\caption{Sequence of points generated by algorithms (a) and (b) with interpolation projection $g$. Since $g$ is not a projection in the $\ell_2$ minimizing sense, it cannot be used as in PGD (a). However, taking the derivative of the projection into account as in (b), drives the algorithm to the optimum.}
	\label{fig:noconv}
\end{figure}

To solve the optimization problem $\min_{x\in {\cal C}}f(x)$ described in \eqref{eq:pbm}, we use a projection $g$ that will ensure that for all $x \in \RR^d$, $g(x) \in \cal C$, i.e. $h(g(x)) \leq 0$. The projection $g$ is defined for any convex function $h$, provided there exists some point $x_0$ strictly satisfying the constraint, i.e. $h(x_0) < 0$. In which case, $g$ is given by 
\[
g(x) = \begin{cases}
x   &\text{if } h(x) \leq 0,\\
\eta_x x + (1-\eta_x) x_0 & \text{else},
\end{cases}
\]
with $\eta_x = \frac{h(x_0)}{h(x_0)-h(x)}$. When $h(x) > 0$, $g$ simply interpolates between the violating point $x$ and the point $x_0$ in $\cal C$; otherwise, it returns $x$ itself.

\begin{proposition}
$g$ is a projection from $\RR^n$ to $\cal C$.\label{eq:prop}
\end{proposition}
\begin{proof}
We will demonstrate that $g(x) \in \cal C$ for all $x \in \RR^d$. If $h(x) \leq 0$, $g(x)$ is in $\cal C$ by definition. If $h(x) > 0$ then  $\eta_x \in (0,1)$ since $h(x_0) - h(x) < h(x_0) < 0$ and 
\begin{align*}
h(g(x)) &= h(\eta_x x + (1-\eta_x) x_0),\\
&\leq \eta_x h(x) + (1-\eta_x) h(x_0), \tag{$h$ convex}\\
&= h(x_0) - \eta_x (h(x_0) - h(x)),\\
&= 0.
\end{align*}
\end{proof}

Even though $g$ is a projection in the sense of Def.~\ref{def:proj}, it is not a projection in the usual sense that it minimizes a norm between $x$ and elements of $\cal C$. As a result, this projection cannot be used as in projected gradient descent (Sec.~\ref{sec:prelim}). To illustrate this, Fig.~\ref{fig:noconv} shows a simple convex problem with a quadratic objective---the sphere function---and a linear constraint. When used as in the projected gradient descent update of Eq.~\eqref{eq:pgd}, the resulting algorithm stales along the line with which it first exits $\cal C$. Indeed, when optimizing the sphere function in an unconstrained way, gradient descent follows a straight line from $x_0$ to the origin. As it first exits $\cal C$, the interpolation projection puts the iterate back on the same line and the algorithm keeps going back and forth indefinitely. In contrast, when optimizing the composition of the projection and the objective by gradient descent
\begin{align}
x_{k+1} = x_k - \alpha_k \nabla f\circ g (x_k),
\label{eq:simpleform}
\end{align}
the iterate is pushed back to $\cal C$ in such a way that it moves towards the optimum. In fact, a simple computation shows us that when $x_k$ is not in $\cal C$, the update in Eq.~\eqref{eq:simpleform} is linearly mixing the gradient of the objective $f$ and the constraint $h$. Formally, when $h(x_k) > 0$, then $g$ is differentiable at $x_k$---from the assumption that $h$ is---and the gradient $\nabla f\circ g (x_k)$ is given by 
\begin{align}
\nabla f\circ g (x_k) &= J_k(x_{k})^T \nabla f(g(x_k)),\notag\\
&= \eta_k \left(I + \frac{\nabla h(x_k)(x_k-x_0)^T}{h(x_0)-h(x_k)}\right) \nabla f(g(x_k)),\notag\\
&= \eta_k \left(\nabla f(g(x_k)) + \frac{\nabla f(g(x_k))^T(g(x_k)-x_0)}{h(x_0)}\nabla h(x_k)\right).\label{eq:grad}
\end{align}
Here $J_k$ is the Jacobian of $g$ at $x_k$, $\eta_k$ is short for $\eta_{x_k}$ and $I$ is the identity matrix. The expression of $J_k$ is obtained by straightforward computation, while Eq.~\eqref{eq:grad} is obtained from the identity $g(x_k)-x_0=\eta_k(x_k - x_0)$. Eq.~\eqref{eq:grad} shows that the gradient of $f\circ g(x_k)$, when $x_k\notin \cal C$, is a linear mixing between the gradient of $f$ at the projected point $g(x_k)$ and the gradient of $h$ at $x_k$. Since $h(x_0) < 0$, the mixing term in Eq.~\eqref{eq:grad} is positive iff $\nabla f(g(x_k))^T(g(x_k)-x_0)\leq 0$. In fact, the first step in our convergence analysis is to show that the previous quantity is indeed always negative. 

The mixing between the gradient of $f$ and $h$ is reminiscent of the conditional subgradient descent of~\cite{larsson1996}. This algorithm is an acceleration of PGD, that restricts the definition of a sub-gradient as a linear under estimator of $f$ only within $\cal C$. In this case, it is shown in~\cite{larsson1996} that when $h(x_k) = 0$, the set of conditional sub-gradients of $f$ can be extended by adding any sub-gradient of $f$ to a sub-gradient of $h$. Here however, the projection $g(x_k)$ is not on the boundary of $\cal C$---for example if $h$ is strictly convex then $h(g(x_k))<0$ and hence Eq.~\eqref{eq:grad} is not necessarily a conditional subgradient of $f$, and the convergence analysis of our algorithm has to be carried out using different tools.
\begin{algorithm}[t]
 \caption{Interpolation-based projection convex optimizer}
 \begin{algorithmic}[1] \label{alg}
  \STATE \textbf{Input:} linear function $f$, convex function $h$, Lipschitz constants $L$ and $H$ (\ref{ass:lin}--\ref{ass:conv}), domain bound $R$ (\ref{ass:dom}), initial point $x_0$ satisfying \ref{ass:feas} and iteration count $K$.
  \STATE Rescale: $h := h/\abs{h(x_0)}$; $H:=H/\abs{h(x_0)}$
  \STATE Step-size: $\beta := \frac{R}{L(1+ HR)\sqrt{K}}$
  \FOR{$k\in \{0,.., K-1\}$}
  \IF{$h(x_k)\leq 0$}
  \STATE $\alpha_k = \beta$
  \STATE $x_{k+1} = x_k -\alpha_k\nabla f(x_k)$
\ELSE  
\STATE $\alpha_k = \abs{h(x_0) - h(x_k)}\beta$
\STATE $x_{k+1} = x_k -\alpha_k\nabla f\circ g(x_k)$ \label{alg:grado}
  \ENDIF
 \ENDFOR
\STATE \textbf{return} $\frac{1}{K}\sum_{k=0}^{K-1}g(x_k)$
 \end{algorithmic}
\end{algorithm}

Alg.~\ref{alg} summarises the optimization algorithm for constrained optimization using the interpolation-based projection. Alg.~\ref{alg} starts by renormalizing $h$ such that $h(x_0) = -1$, then defines the optimal step-size $\beta$ w.r.t. an upper bound derived given assumptions \ref{ass:lin} to \ref{ass:dom} defined in the next section. Alg.~\ref{alg} then follows a gradient descent (Eq.~\eqref{eq:simpleform}), selecting a different step-size $\alpha_k$, as a function of a constant $\beta$, whether the iterate is inside or outside $\cal C$. When $x \notin \cal C$, the gradient is given by Eq.~\eqref{eq:grad}. Alg.~\ref{alg} then returns the average of the projected points. The algorithm operates a first order gradient descent on $f\circ g$, which as per Eq.~\eqref{eq:grad}, is of linear time and memory complexity.

\section{Convergence analysis}
\label{sec:conv}
The first step in the convergence analysis of Alg.~\ref{alg} is a lemma showing that for an appropriate choice of the step-size $\alpha_k$, the quantity $\nabla f(g(x_k))^T(g(x_k)-x_0)$ is always negative for $k \geq 0$. As a consequence, the gradient of $f\circ g$ will always mix gradients of objective and constraint with opposing directions when the iterate exits the valid set. We prove the lemma under the assumption of a linear objective function $f$, a Lipschitz continuous domain defining function $h$, in addition to the previously discussed assumption of an initial strictly feasible point $x_0$. 

\begin{enumerate}[leftmargin=1cm, label=\textbf{A\arabic*.}, ref=A\arabic*, series=assump]
\item $f(x) = c^T x$ is a linear function in $\RR^d$ and $\norm{c}_2 \leq L$.\label{ass:lin}
\item 	$h$ is convex, everywhere differentiable in $\RR^d$ and H-Lipschitz w.r.t. $\norm{.}_2$. \label{ass:conv}
\item There exists $x_0$ such that $h(x_0) < 0$. \label{ass:feas}
\end{enumerate}

\begin{lemma}
Under \ref{ass:lin}--\ref{ass:feas}, the sequence of $x_k$ produced by Alg.~\ref{alg} verifies, for all $k\geq 0$ and for $\beta \leq \frac{1}{LH}$, $\nabla f(g(x_k))^T(g(x_k)-x_0) \leq 0$.
\label{lem:neg}
\end{lemma}

\begin{proof}
	Let us prove the lemma by induction. For $k = 0$ the inequality is trivially true. Now assuming the inequality holds for some $k \geq 0$. It implies that $c^T(g(x_k) - x_0) \leq 0$. We distinguish in the following two cases, whether $x_k$ is feasible or not. However, we treat both cases of feasibility of $x_{k+1}$ jointly by writing $g(x_{k+1}) - x_0 = \eta_{k+1}\left(x_{k+1}-x_0\right)$ which becomes true by assuming $\eta_{k+1} = 1$ when $x_{k+1}$ is feasible.  
	 First, assume $h(x_k) \leq 0$ then
	\begin{align*}
      \nabla f(g(x_{k+1}))^T(g(x_{k+1})-x_0) &= \eta_{k+1}c^T(x_{k+1}-x_0).
	\end{align*}
	By adding and subtracting $x_k$ inside the parentheses, and since for $h(x_k)\leq 0$, $x_{k+1}-x_k = -\alpha_k c$, we arrive at
	\begin{align*}
	\nabla f(g(x_{k+1}))^T(g(x_{k+1})-x_0)&= \eta_{k+1}\big(-\alpha_k c^Tc +c^T (x_k-x_0)\big),
	\end{align*}	 
	which from the induction hypothesis is the sum of two negative numbers and is thus negative. Now if $h(x_k) > 0$  then by again adding and subtracting $x_k$, and by replacing $x_{k+1}-x_k$ with the gradient update following Eq.~\eqref{eq:grad}, we obtain 
	\begin{align*}
		\nabla f(g(x_{k+1}))^T(g(x_{k+1})-x_0)= \eta_{k+1}\Bigg(-&\alpha_k \eta_k c^Tc + c^T (x_k-x_0)\\
		&\bigg(1 - \frac{\alpha_k \eta_k}{h(x_0)-h(x_k)}c^T\nabla h(x_k)\bigg)\Bigg).
	\end{align*}
	From the induction hypothesis, it is sufficient that $\frac{\alpha_k  \eta_k}{h(x_0)-h(x_k)}c^T\nabla h(x_k) \leq 1$ for the last quantity to be negative.
	Using the fact that \begin{align*}
		\frac{\alpha_k  \eta_k}{h(x_0)-h(x_k)}c^T\nabla h(x_k) &\leq \left|\frac{\alpha_k  \eta_k}{h(x_0)-h(x_k)}c^T\nabla h(x_k)\right|,
\end{align*}
and using the Cauchy-Schwarz inequality as well as assumption \ref{ass:lin} and \ref{ass:conv}, we obtain  
\begin{align*}
		\frac{\alpha_k  \eta_k}{h(x_0)-h(x_k)}c^T\nabla h(x_k)
		&\leq \left|\frac{\alpha_k\eta_k}{h(x_0)-h(x_k)}\right|LH, \\
		&\leq \beta LH,\tag{$\eta_k < 1$}\\
	\end{align*}
	Since $\beta \leq \frac{1}{LH}$ by assumption, the last quantity is $\leq 1$ as desired. As such, we conclude that $\nabla f(g(x_{k+1}))^T(g(x_{k+1})-x_0) \leq 0$ for $h(x_k) > 0$.
\end{proof}

The assumption of the linearity of $f$ is used in the induction step and allows several simplifications since for $f$ linear, $\nabla f(x_{k+1}) = \nabla f(x_k)$. Extending the convergence analysis of Alg.~\ref{alg} to non-linear objectives could be achieved by extending Lem.~\ref{lem:neg} to this case. However, as discussed in Sec.~\ref{sec:nonlin}, since the assumptions on $h$ are mild, many constrained convex optimization algorithms can be recast as problems solvable by Alg.~\ref{alg}.

To prove convergence of Alg.~\ref{alg}, we need an additional assumption on the boundedness of the distance to an optimum. 

\begin{enumerate}[leftmargin=1cm, label=\textbf{A\arabic*.}, ref=A\arabic*, resume=assump]
\item $\exists x^* \in {\cal C}$ such that $\forall x \in {\cal C}, f(x^*)\leq f(x)$ and $\norm{x_0 - x^*} \leq R$, for some $R \geq 0$. \label{ass:dom}
\end{enumerate}

The convergence result for Alg.~\ref{alg} is as follows
\begin{theorem}
\label{thm}
Under \ref{ass:lin}--\ref{ass:dom} and for $H_0 = \frac{H}{\abs{h(x_0)}}$, the returned value of Alg.~\ref{alg} verifies $f\left(\frac{1}{K}\sum_{k=0}^{K-1} g(x_k)\right) - f(x^*) \leq \frac{RL(1+ H_0R)}{\sqrt{K}}$ for $K \geq \frac{R^2H_0^2}{(1+ H_0R)^2}$ and for $\beta = \frac{R}{L(1+ H_0R)\sqrt{K}}$.
\end{theorem}

\begin{proof}
As \ref{ass:feas} ensures that $h(x_0)$ is non zero, an equivalent optimization problem can be obtained where $h(x_0) = -1$ by rescaling $h$ with $\abs{h(x_0)}$. Letting $H_0 = \frac{H}{\abs{h(x_0)}}$, the only difference will be that if $h$ is $H$-Lipschitz then $h/\abs{h(x_0)}$ is $H_0$-Lipschitz. From now on, and without loss of generality, we assume that $h(x_0) = -1$ and $h$ is $H$-Lipschitz. We revert to the general case where $h(x_0) < 0$ at the end of the proof.

Following standard proofs of subgradient descent algorithms, our proof begins by estimating the distance of the iterate to the optimum
\begin{align*}
\norm{x_{k+1}-x^*}_2^2 &= \norm{x_k - \alpha_k \nabla f\circ g (x_k) - x^*}_2^2.\\
\end{align*}
As in Lem.~\ref{lem:neg}, we study separately the case where $x_k\in \cal C$ and $x_k\notin\cal C$. In each case, we derive an upper bound of $\norm{x_{k+1}-x^*}_2^2$ and then pick the largest of the two. Starting with $x_k\notin\cal C$, we replace $ \nabla f\circ g (x_k)$ by its definition in Eq.~\eqref{eq:grad}, and by expanding the quadratic expression we obtain
\begin{align}
\norm{x_{k+1}-x^*}_2^2 &= \norm{x_k-x^*}_2^2 + \norm{\alpha_k \nabla f\circ g (x_k)}_2^2 - 2\alpha_k \eta_k \nabla  f(g(x_k))^T(x_k-x^*)\notag\\
&\quad - 2\alpha_k \eta_k \frac{ \nabla f(g(x_k))^T(x_k-x_0)\nabla h(x_k)^T(x_k-x^*)}{h(x_0)-h(x_k)}.\label{eq:1th1}
\end{align}
Adding and subtracting $g(x_k)$ in $\nabla  f(g(x_k))^T(x_k-x^*)$ and by expanding the definition of $g(x_k)$ and $\eta_k$ when $h(x_k) > 0$ we obtain
\begin{align*}
\nabla  f(g(x_k))^T(x_k-x^*) = \nabla f(g(x_k))^T(&g(x_k)-x^*) \\
&-\frac{h(x_k)}{h(x_0)-h(x_k)}\nabla f(g(x_k))^T(x_k - x_0).
\end{align*}
Replacing $\nabla  f(g(x_k))^T(x_k-x^*)$ in Eq.~\eqref{eq:1th1} gives
\begin{align}
&\norm{x_{k+1}-x^*}_2^2 = \norm{x_k-x^*}_2^2 + \norm{\alpha_k \nabla f\circ g (x_k)}_2^2 - 2\alpha_k \eta_k \nabla f(g(x_k))^T(g(x_k)-x^*)\notag\\
&\qquad\qquad\ + 2\alpha_k \eta_k \left(\frac{h(x_k) + \nabla h(x_k)^T(x^*-x_k)}{h(x_0)}\right) \nabla f(g(x_k))^T(g(x_k)-x_0).\label{eq:1th2}
\end{align}
But from convexity of $h$, we know that $h(x_k) + \nabla h(x_k)^T(x^*-x_k) \leq h(x^*) \leq 0$ implying 
\begin{align*}
\frac{h(x_k) + \nabla h(x_k)^T(x^*-x_k)}{h(x_0)}\geq \frac{h(x^*)}{h(x_0)}\geq 0.
\end{align*}
In addition, $\alpha_k$ and $\eta_k$ are always positive and from Lem.~\ref{lem:neg}, $\nabla f(g(x_k))^T(g(x_k)-x_0)$ is negative for all $k\geq 0$ provided $\beta \leq \frac{1}{LH}$. As a result the last term of Eq.~\eqref{eq:1th2} is always negative and $\norm{x_{k+1}-x^*}_2^2$ can be bounded by
\begin{align}
\norm{x_{k+1}-x^*}_2^2 \leq \norm{x_k-x^*}_2^2 + &\norm{\alpha_k \nabla f\circ g (x_k)}_2^2\notag \\
&\qquad- 2\alpha_k \eta_k \nabla f(g(x_k))^T(g(x_k)-x^*).\label{eq:1th3}
\end{align}
In the upper bound of Inq.~\eqref{eq:1th3}, we will now bound the term $\norm{\alpha_k \nabla f\circ g (x_k)}_2^2$ that is specific to the case $h(x_k) > 0$. By replacing the gradient with its definition and using the fact that we have rescaled $h$ such that $h(x) = -1$, we obtain
\begin{align*}
\beta^{-2}||\alpha_k \nabla &f\circ g (x_k)||_2^2
=||\nabla f(g(x_k)) - \nabla f(g(x_k))^T(g(x_k)-x_0)\nabla h(x_k)||_2^2.\\
\end{align*}
Using the Cauchy-Schwarz inequality as well as assumption \ref{ass:lin}, \ref{ass:conv} and \ref{ass:dom} we obtain
\begin{align}
\beta^{-2}\norm{\alpha_k \nabla f\circ g (x_k)}_2^2 \leq L^2 (1+ HR)^2.\label{eq:1th_norm}
\end{align}
Replacing Eq.~\eqref{eq:1th_norm} into Eq.~\eqref{eq:1th3}, using the definition of $\alpha_k$ and since $h(x_0) = -1$ we have
\begin{align}
\norm{x_{k+1}-x^*}_2^2 \leq &\norm{x_k-x^*}_2^2 + \beta^2 L^2 (1+ HR)^2 - 2\beta \nabla f(g(x_k))^T(g(x_k)-x^*).\label{eq:1th_out}
\end{align}
Now for the simpler case $x_k \in \cal C$ we have
\begin{align*}
\norm{x_{k+1}-x^*}_2^2 &= \norm{x_{k}-x^*}_2^2 + \norm{\alpha_k\nabla f(x_k)}_2^2-2\alpha_k\nabla f(x_k)^T(x_k-x^*),\\
\end{align*}
Using assumption \ref{ass:lin} and since $x_k = g(x_k)$ and $\alpha_k = \beta$ when $x_k \in \cal C$, we obtain the following bound
\begin{align}
\norm{x_{k+1}-x^*}_2^2 \leq &\norm{x_{k}-x^*}_2^2 + \beta^2L^2 -2\beta\nabla f(g(x_k))^T(g(x_k)-x^*).\label{eq:1th4}
\end{align}
Clearly the upper bound of $\norm{x_{k+1}-x^*}_2^2$ in Inq.~\eqref{eq:1th_out} is always larger than the one in Inq.~\eqref{eq:1th4}. As such, we can use the upper bound of $\norm{x_{k+1}-x^*}_2^2$ in Inq.~\eqref{eq:1th_out} for all iterates of Alg.~\ref{alg}. Letting $A = L^2 (1+ HR)^2$, and averaging over the first $K$ terms of both sides of Inq.~\eqref{eq:1th4} yields
\begin{align*}
\frac{1}{K}\sum_{k=0}^{K-1}\norm{x_{k+1}-x^*}_2^2 \leq \frac{1}{K}\sum_{k=0}^{K-1}\norm{x_k-x^*}_2^2 &+ \beta^2 A \\
&- \frac{2\beta}{K} \sum_{k=0}^{K-1}\nabla f(g(x_k))^T(g(x_k)-x^*).\\
\end{align*}
From the convexity of $f$ we have that
\begin{align*}
 \nabla f(g(x_k))^T(g(x_k)-x^*)&\geq   f(g(x_k))-f(x^*),
\end{align*}
as well as $\frac{1}{K}\sum_{k=0}^{K-1} f(g(x_k))\geq  f\left(\frac{1}{K}\sum_{k=0}^{K-1}g(x_k)\right)$. Using these two properties yields 
\begin{align*}
\frac{1}{K}\sum_{k=0}^{K-1}\norm{x_{k+1}-x^*} _2^2\leq  \frac{1}{K}\sum_{k=0}^{K-1}\norm{x_k-x^*}_2^2 &+ \beta^2 A \\
&- 2\beta  \left(f\left(\frac{1}{K}\sum_{k=0}^{K-1}g(x_k)\right)-f(x^*)\right)
\end{align*}
Rearranging terms and cancelling telescoping sums yields
\begin{align*}
f\left(\frac{1}{K}\sum_{k=0}^{K-1}g(x_k)\right)-f(x^*) &\leq \frac{1}{2\beta K}\Big(\norm{x_0 - x^*}_2^2 -\norm{x_{K} - x^*}_2^2 + K\beta^2A\Big).\notag\\
\end{align*}
Using \ref{ass:lin}, \ref{ass:conv} and \ref{ass:dom} and after replacing $A$ we obtain
\begin{align*}
f\left(\frac{1}{K}\sum_{k=0}^{K-1}g(x_k)\right)-f(x^*)\leq \frac{R^2}{2\beta K}+\frac{\beta L^2(1+ HR)^2}{2}.
\end{align*}
Minimizing this upper bound w.r.t. to $\beta$ gives the optimal fixed step-size $\beta = \frac{R}{L(1+ HR)\sqrt{K}}$ with error
\begin{align}
f\left(\frac{1}{K}\sum_{k=0}^{K-1}g(x_k)\right)-f(x^*) 
&\leq \frac{RL(1+ HR)}{\sqrt{K}}.\label{eq:boundnoh}
\end{align}
This gives us a first condition on $\beta$, but to achieve the bound in Inq.~\eqref{eq:boundnoh}, we made use of Lem.~\ref{lem:neg} which requires that $\beta \leq \frac{1}{LH}$, yielding an additional condition on $K$
\begin{align}
&\quad\quad\frac{R}{L(1+ HR)\sqrt{K}} \leq \frac{1}{LH}\notag,\\
&\Leftrightarrow K \geq \frac{R^2H^2}{(1+ HR)^2}.\label{eq:condK}
\end{align}
Now the only remaining operation is to express the step-size, the condition on $K$ in Inq.~\eqref{eq:condK} and the error upper bound in Inq.~\eqref{eq:boundnoh} in terms of the original Lipschitz constant which is achieved simply by replacing $H$ with $\frac{H}{\abs{h(x_0)}}$ in these inequalities.
\end{proof}

The $\mathcal{O}(\frac{1}{\sqrt{K}})$ convergence rate is typical of sub-gradient descent on non-smooth convex functions \citep{nocedal2006}, which is expected since $f\circ g$ is non-smooth. Compared to projected gradient descent (PGD), the bound now shows an explicit dependence on the Lipschitz constant of $h$. This is also expected since in PGD the projection is assumed to be computable at no cost. As a result, the error bound of PGD does not depend on the gradient of $h$ in any way, whereas in our algorithm this dependence is made explicit. Because of the non-smoothness of $f\circ g$ and the resulting $\mathcal{O}(\frac{1}{\sqrt{K}})$ convergence rate, we do not expect the general formulation of Alg.~\ref{alg} to be competitive with specialized convex optimizers developed for specific convex problem classes. However, the versatility and cheap computational cost of the interpolation projection offers large gains compared to convex optimizers when integrated into (non-convex) machine learning models, as shown in the experimental validation section.

\subsection{Subgradients, multiple constraints and non-linear objectives}
\label{sec:sub}
\label{sec:nonlin}
So far we have only considered a single inequality constraint. Alg.~\ref{alg} and its theoretical guaranties can easily be extended to tackle multiple inequality constraints and an affine equality constraint
\begin{equation*}
\begin{aligned}
\min_{x \in \RR^d} \quad &	 f(x),\\
\text{s.t.} \quad & h_i(x)\leq 0, \ \text{for all } i \in \{1\dots M\},\\
& Ax = b,
\end{aligned}
\end{equation*}
where $h_i$ are convex functions in $\RR^d$,  $A$ a matrix and $b$ a vector. Let ${\cal C'}=\{x\in\RR^d: h_i(x) \leq 0\, \text{for all } i \in \{1\dots M\}\}$. We define $h$ as $h(x) = \max_{i\in \{1\dots M\}} h_i(x)$.
Then $h$ is sub-differentiable if all $h_i$ are (sub-)differentiable. Moreover, we assume that all $h_i$ are Lipschitz with constant at most $H$, resulting in the following assumption
\begin{enumerate}[leftmargin=1cm, label=\textbf{A\arabic*.}, ref=A\arabic*, resume=assump]
\item $h$ is convex, sub-differentiable in $\RR^d$ and H-Lipschitz w.r.t. $\norm{.}_2$. \label{ass:subdiff}
\end{enumerate}
To tackle constrained optimization in $\cal C'$, we define Alg.~\ref{alg}' that replaces Line 10 of Alg.~\ref{alg}. Specifically, the gradient $\nabla h$ in Eq.~\eqref{eq:grad} is simply replaced by a sub-gradient of $h$. Under \ref{ass:lin}, \ref{ass:feas}--\ref{ass:subdiff}, this new algorithm has the same convergence properties of Alg.~\ref{alg}. Indeed, $h$ being convex, the projection is still valid and will be given with interpolation weight $\eta_x = \min_{i \in \{1\dots M\}}\frac{h(x_0)}{h(x_0)-h_i(x)}$, selecting the smallest interpolation weight given by the constraint $h_i$ with the highest violation. Additionally, of $h$, the proof of Thm.~\ref{thm} only uses the property $\nabla h(x_k)^T(x^*-x_k) \leq h(x^*)-h(x_k)$ which is also fulfilled by a sub-gradient of $h$.

In summary, the differentiablity requirement of $h$ can be relaxed to only require sub-differentiability, and multiple constraints are treated as a single constraint using the $\max$ over these sub-differentiable constraints. As for the affine equality constraint, it can be eliminated by replacing $x$ with $Fz + x_0$ as shown in~\cite{boyd2004}, where $F$ is a matrix whose range is the null space of $A$ under the condition that $x_0$ is a solution of $Ax = b$. Note that the objective function remains linear after the aforementioned change of variable, and hence the convergence guarantees still apply.

As for non-linear objectives, we note that most convex programs can be written as cone programs of the form $\min_{x \in \mathcal{K}} c^T x$, for a closed convex cone $K$ and a linear objective~\citep{nesterov1994}. In fact, there exists automated tools \citep{grant2006, grant2008} that perform this rewriting by replacing non-linear functions in the computational graph with their graph implementation---a generic epigraph-based representation. These tools are used by existing solvers such as CVX \citep{cvx}, and for our algorithm to be applicable to these cone programs, one has to provide a domain defining function $h$ equivalent to the constraint ${x \in \mathcal{K}}$ for all cones supported by the tool. In the next section, we provide numerical examples for the semi-definite cone, the second order cone and the linear cone. 

\section{Experimental validation}
\label{sec:num}
We first conduct numerical evaluations on toy convex problems to validate the theoretical analysis. The broader usage of the interpolation projection in machine learning is then evaluated in both a reinforcement and supervised learning setting.
\subsection{Constrained convex optimization}
Alg.~\ref{alg} defines the step-size as a function of the domain bounds and the Lipschitz constants which are typically unknown in practice. We thus investigate on a wide range of convex optimization problems the robustness of the interpolation projection to the choice of (a potentially wrong) step-size. We compare our algorithm to Projected Gradient Descent (PGD, ~\cite{rosen1960,nocedal2006}) and subgradient descent (SubGD,~\cite{shor1985,bertsekas2015}). Subgradient descent is a converging descent algorithm that in our constrained setting operates by i) following the gradient of $f$ if $x \in \cal C$ ii) following the (sub-)gradient of $h$ otherwise. This algorithm is very simple and another objective of these numerical experiments is to investigate whether the mixing of the gradients $\nabla f$ and $\nabla h$, obtained from differentiating through $f\circ g$ in Eq.~\eqref{eq:grad}, provides any practical advantage compared to the simpler scheme of subgradient descent. In the following, we denote our algorithm by IGD, where the `I' stands for interpolation. We consider five problem classes comprising linear programs, semi-definite programs, second order cone programs, problems with a bounded $\ell_2$ norm or with an exponential form constraint. Exact definition of each problem and their random generation process is deferred to the appendix. 

\textbf{Results.}  For each of the five problem classes, 100 random instances are generated and we compute at each iteration the smallest $\frac{f(x_k) - f(x^*)}{f(x_0) - f(x^*)}$ achieved so far. We compared the gradient descent algorithms with step-sizes from $10^{-4}$ to $10^{-1}$. Experiments for each step-size are conducted on the same 100 problem instances, and although we plot the results for each step-size separately, one can easily extract the best performing step-size for each method from the same plots. The plots~(deferred to the appendix) show that in 17 out of the 20 problems and step-sizes combinations, IGD outperforms SubGD, sometimes with several order of magnitudes. On semi-definite programs, SubGD performs better with larger step-sizes, although best results are still obtained overall by IGD with the smallest step-size. On the bounded norm problem where PGD is applicable, our algorithm is able to match PGD up until a precision ranging from $10^{-2}$ to $10^{-5}$ depending on the step-size, before tracking behind. In contrast, SubGD is distanced at a significantly lower precision. These results both demonstrate a certain robustness to the choice of step-size and a practical interest in the mixing of gradients obtained by differentiating through $f\circ g$. Thanks to the generality of the projection and the simplicity of performing unconstrained gradient descent on $f\circ g$, we expect the interpolation projection to find many usages in machine learning, two of which are presented in the next subsections.

\subsection{Reinforcement learning in continuous action spaces}
\label{sec:rl}
We consider in this section policy optimization updates that occur at each iteration of the approximate policy iteration (API) scheme \citep{Bertsekas11, Scherrer14}. To formalize the policy update in API we briefly introduce key concepts of reinforcement learning (RL). A Markov Decision Process (MDP) is a quintuple $(\State, \Action, R, P, \gamma)$ where $\State$ and $\Action$ are state and action spaces, that are in our experiment  $\RR^{d_s}$ and $\RR^{d_a}$ respectively. $P: \State \times \Action \mapsto {\cal P}(\State)$ and $R: \State \times \Action \mapsto \R$ determine the next state transition probability and reward upon the execution of a given action in a given state. We denote by $q(a| s)$ the probability density of executing $a \in \Action$ in $s \in \State$ according to the stochastic policy $q$. 
Additionally, for policy $q$ we define the Q-function $Q_{q}(s,a) = \EE\left[\sum_{t=0}^\infty\gamma^t R(s_t, a_t)\mid s_0 = s, a_0=a\right]$,  where the expectation is taken w.r.t. random variables $a_{t+1}\sim q(.|s_t)$ and $s_{t+1}\sim p(.|s_t, a_t)$ for $t > 0$; the value function $V_{q}(s) = \EE_{a\sim q(.|s)}\left[Q_{q}(s,a)\right]$ and the advantage function $A_{q}(s,a)=Q_{q}(s,a) - V_{q}(s)$. The goal in API is to find the policy maximizing the policy return $J(q) = V_{q}(s_0)$ for some starting state $s_0$. 

API iterates three steps, generating data from the current policy $q$, evaluating $A_q$ and updating the policy $q$ using $A_q$. To update the policy we consider the maximization of $A_q$ under a KL divergence constraint between the current and next policies---establishing a 'step-size' in probability space---as is done in \cite{Schulman15, rajeswaran17, peters08}. The policy  update is given by 
\begin{alignat}{3}
& \underset{p}{\arg\max} && \EE_{s,a\sim q}\left[\frac{p(a|s)}{q(a|s)}A_q(s,a)\right] \label{eq:objapi}\\
& \text{subject to}\ \ \ \ \ 
&& \EE_{s \sim q}\left[\KLM{p(.|s)}{q(.|s)}\right] &&\leq \epsilon.\label{eq:KLapi}
\end{alignat}
We specifically consider the setting where $p$ and $q$ are Gaussian policies. A Gaussian policy has density $p(.|s) = \Normal(\mu(s), \Sigma)$, for co-variance matrix $\Sigma$ and mean function $\mu(.)$. In our set-up we consider diagonal co-variance matrices as in \cite{Schulman15, rajeswaran17} and linear-in-features or neural network based mean functions. The linear-in-feature mean function is given by $\mu(s) = \phi(s)^TM$ using the same random Fourier features $\phi$ of \cite{rajeswaran17} with 2000 entries, whereas the neural network mean function is given by a neural network following the architecture in \cite{Schulman15} with 2 hidden layers with 64 neurons each. For estimating $A_q$ we follow \cite{rajeswaran17} and use a neural network to learn $V_q$ and estimate $A_q$ from trajectories. For both cases we use $\epsilon = 10^{-2}$ as in \cite{Schulman15}.


To solve the aforementioned problems, both natural approaches with linear-in-features \citep{rajeswaran17} and neural network mean functions \citep{Schulman15} follow the same approach: a second order approximation of the constraint \eqref{eq:KLapi} is computed, as well as a linear approximation of the objective function \eqref{eq:objapi}. The resulting problem is then solved in closed form resulting in the natural gradient update of the policy parameters. However, as the constraint satisfaction is not guaranteed---since the problem is solved by approximating the constraint---both approaches \citep{Schulman15, rajeswaran17} add a line-search routine, interpolating between the new parameters and the parameters of $q$, to ensure that Inq.~\eqref{eq:KLapi} holds.

\begin{figure*}
	\centering
	\def\figBenefitsWidth{.26}
	\begin{subfigure}[h]{\figBenefitsWidth\textwidth}        
		\includegraphics[width=\linewidth]{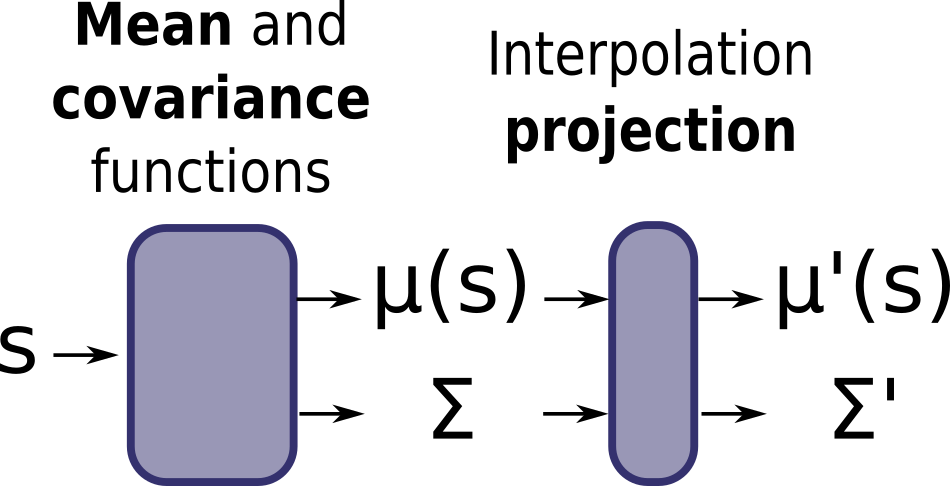}
		
	\end{subfigure}
	\begin{subfigure}[h]{\figBenefitsWidth\textwidth}        
		\includegraphics[width=\linewidth]{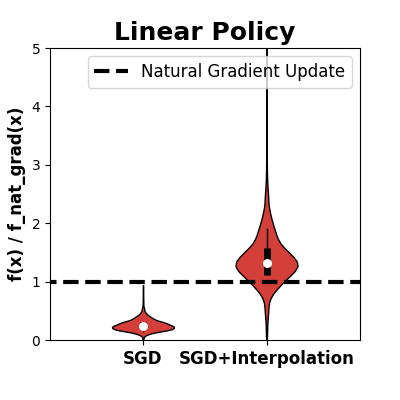}       
		
	\end{subfigure}
	\begin{subfigure}[h]{\figBenefitsWidth\textwidth}        
		\includegraphics[width=\linewidth]{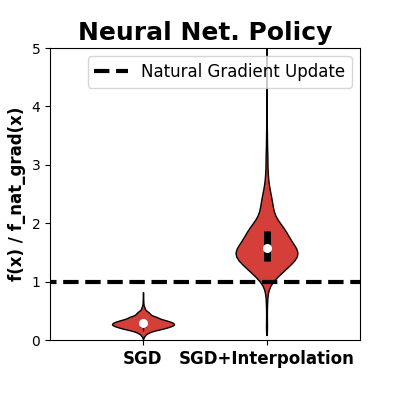}
	\end{subfigure}
	\caption{From left to right: a) The computational graph of an RL policy with the projection layer taking as input the intermediate values $\mu(s)$ and $\Sigma$ and returning a new mean and covariance complying with the KL-divergence constraint. b-c) Distributions of the improvement ratio over the natural gradient baseline for gradient descent on the policy parameters with and without the interpolation projection. The thick vertical black bars in the violin plot span the lower and upper quartiles.} 
	\label{fig:rl}
	\vspace*{-.2in}
\end{figure*}
To compare to natural gradient, we employ first a naive algorithm that optimizes objective \eqref{eq:objapi} in an unconstrained way, with the Adam algorithm~\citep{KingmaB14}, before calling the line-search routine used by the natural gradient approaches to ensure constraint satisfaction. 
Secondly, we augment the naive algorithm by adding an interpolation projection 'layer' to the output of the policy. The projection layer, as depicted in Fig. 2-left, takes as input a set of action means---given by evaluating the current mean function over a mini-batch of input states---and a covariance matrix and returns a new set of means and a covariance matrix that comply with the constraint. To formalize, let us define $h$ and $x_0$, the two elements needed to perform the interpolation projection. Given a finite set of states $\{s_1, \dots, s_K\}$, we define 
\[
h(\mu(s_1),\dots,\mu(s_K), \Sigma) = \frac{1}{K}\sum_{k}\text{KL}(\Normal(\mu(s_k),\Sigma)|\Normal(\mu_q(s_k), \Sigma_q)) - \epsilon,
\]
where $\mu_q$ and $\Sigma_q$ are respectively the mean function and covariance matrix of $q$. $h$ is convex and we use as $x_0$ for the interpolation projection the means and covariance matrix of $q$. The projection that returns a set of means and a covariance matrix compying with the KL divergence constraint is then given by $g$ as in Sec.~\ref{sec:alg}, from the definition of $h$ and $x_0$.

To illustrate the algorithm, assume for a mini-batch of states $\{s_1, \dots, s_K\}$ the mean and covariance functions return a mini-batch of means $\mu(s_1),\dots,\mu(s_K)$ and a covariance matrix $\Sigma$. If the constraint, estimated for this mini-batch is violated,
\[
\frac{1}{K}\sum_{k}\text{KL}(\Normal(\mu(s_k),\Sigma)|\Normal(\mu_q(s_k), \Sigma_q)) > \epsilon,
\]
we use the projection $g$ as in Sec.~\ref{sec:alg} to obtain a new set of means $\mu_\eta(s_1),\dots,\mu_\eta(s_K)$ and covariance matrix $\Sigma_\eta$ where $\mu_\eta(s_k) = \eta \mu(s_k) + (1-\eta) \mu_q(s)$ and $\Sigma_\eta = \eta \Sigma + (1-\eta) \Sigma_q$ and then evaluate the objective for $p_\eta$
\[
\frac{1}{N}\sum_{k}\frac{p_\eta(a_k|s_k)}{q(a_k|s_k)}A_q(s_k,a_k),
\] 
where $p_\eta(.|s) = \Normal(\mu_\eta(s),\Sigma_\eta)$. Once the objective is computed, we backpropagate throughout the whole computational graph which backpropagates through the interpolation projection. 

In the linear-in-feature case, we note that the KL divergence is not only convex in the mean and covariance of the Gaussian but also in the policy parameters. Specifically, we have that 
\[
h(M, \Sigma) = \frac{1}{N}\sum_{k}\text{KL}(\Normal(\phi(s_k)^TM,\Sigma)|\Normal(\phi(s_k)^TM_q, \Sigma_q)) - \epsilon,
\]
is a convex function in $M$ and $\Sigma$, and from linearity of the mean function interpolating the means or the parameter $M$ directly are equivalent. Moreover, the $\eta$ obtained using $h(M, \Sigma)$ or $h(\mu(s_1),\dots,\mu(s_K), \Sigma)$ will be identical for a given mini-batch since the value of $h$ will be the same in both cases. The optimization process can thus be seen as performing gradient descent on $(f\circ g)(M, \Sigma)$, where $f$ is the objective \eqref{eq:objapi}. This is similar to the convex optimization setting studied theoretically, except $f$ is now non-linear non-convex---because $A_q$ is not necessarily convex. However, the empirical results show that the optimization scheme still performs well despite $f\circ g$ being non-convex. This is not entirely surprising since gradient descent is widely used and well behaved for non-convex problems too. 

\begin{figure*}
\centering
\includegraphics[width=.3\textwidth]{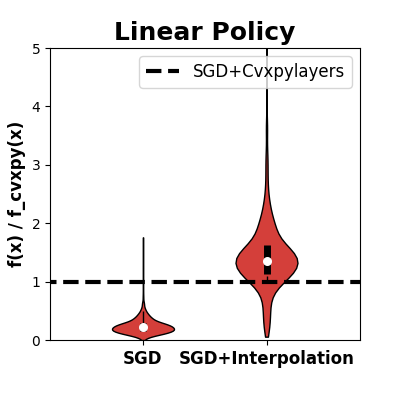}
\includegraphics[width=.3\textwidth]{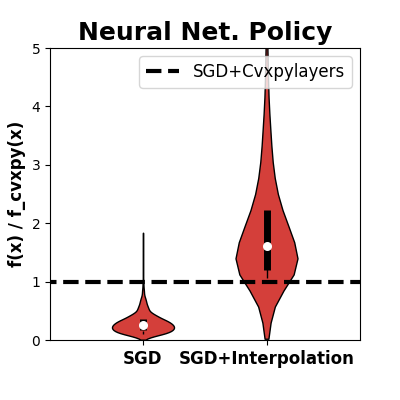}
\caption{Distributions of the improvement ratio over SGD + A norm minimizing projection of SGD with and without the interpolation projection. The thick black bars in the violin plot span the lower and upper quartiles. Each violin plot is obtained after solving circa 1700 optimization problems.}
\label{fig:cvx}
\end{figure*}

To generate real RL optimization problems, we run natural gradient on the \texttt{BipedalWalker-v2} environment \citep{gym} for one million steps with a policy update after a minimum of 3000 steps. We run 11 of such independent runs, generating over 3000 optimization problems for each of the linear and non-linear cases. Both the naive algorithm and the projection augmented algorithm use the same hyper-parameters for the update, by performing 30 epochs with a step-size\footnote{We performed the same experiment with other step-sizes of $10^{-4}$ and $2\times 10^{-4}$ and the conclusions are essentially the same.} of $5\times 10^{-5}$). For each of the 3000 optimization problems, we record the ratio between the objective value when solving the problem with gradient descent, divided by the value when solving the problem following the natural gradient baselines in each of the linear \citep{rajeswaran17} and non-linear \citep{Schulman15} case. A value larger than 1 indicates that the method solved the constrained problem better than the state-of-the-art. 

Fig.~\ref{fig:rl} shows the distribution of such ratios for the linear and non-linear mean function cases. In both cases, without the projection, the unconstrained optimization with a final line-search step performs significantly worse than natural gradient descent. In contrast, adding the interpolation projection of the Gaussian distributions' parameters while using the same optimization scheme, results in a median improvement over natural gradient of $31\%$ and $57\%$ for the linear and non-linear mean function cases respectively. Note that in the linear case, the optimization setting resembles the earlier convex optimization experiments as the constraint is convex in the input means of $h$ but also directly on the parameters of the mean function $M$. When the mean function is a neural network, the interpolation projection still seems to guide the gradient descent algorithm towards regions of the parameter space that better trade off objective maximization and constraint satisfaction than the naive algorithm.

We also evaluated replacing the interpolation layer with an orthogonal projection using a differentiable convex solver \citep{agrawal2019}. The orthogonal projection receives the same input means and covariance matrix as the interpolation projection but returns instead the parameters that minimize the Euclidean distance to the inputs while complying with the KL divergence constraint. This is a convex problem and we used the tools of \citep{agrawal2019} to both compute the forward pass---solve the convex problem---and the backward pass---differentiate around the solution of the convex problem---of this computational graph. The computational cost of this model is more than 300 times that of the vanilla neural network model, while our model with the interpolation projection is only about 1.5 more expensive. Due to the increased computational costs, we performed only 6 independent runs for this comparison totaling about 1700 optimization problems. Comparison between the two optimization schemes are shown in Fig.~\ref{fig:cvx}. Surprisingly, the interpolation projection performs better than the more accurate projection, perhaps because of a better interplay between the interpolation projection and the subsequent line-search routine, while being significantly cheaper to compute.
\subsection{Supervised learning of dynamics models}
\begin{figure*}
\begin{subfigure}[b]{.38\textwidth}
\centering
\includegraphics[width=\textwidth]{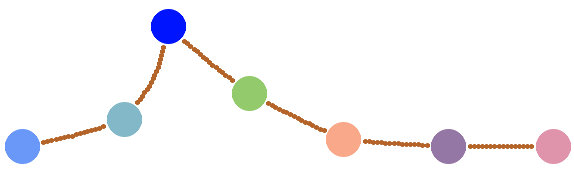}
\caption{Chain}
\end{subfigure}
\begin{subfigure}[b]{.38\textwidth}
\centering
\includegraphics[height=.7in]{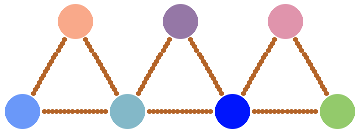}
\caption{Chained triangles}
\end{subfigure}
\begin{subfigure}[b]{.20\textwidth}
\centering
\includegraphics[height=.75in]{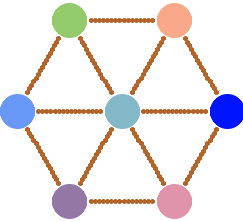}
\caption{Star}
\end{subfigure}
\caption{The three considered objects with 7 rigid bodies and 6, 9 and 12 strings respectively from left to right.}
\label{fig:shapes}
\end{figure*}
\begin{table}[]
    \centering
    \begin{tabular}{l|c|c|c}
       & \textbf{Chain} &\textbf{ Chain. Tri.} & \textbf{Star} \\ 
        \hline\hline
        \textbf{RNN} & $\mathbf{2.52 \pm 1.38}$  & $2.32 \pm 1.19$ & $2.25\pm 1.09$ \\
        \hline
        \textbf{RNN + Shape Cst.} & $2.89 \pm 1.39$ & $\mathbf{2.19\pm 1.01}$ & $\mathbf{2.18\pm 0.96}$
    \end{tabular}
    \caption{Mean Euclidean distance and std. dev. between test trajectories and model generated trajectories,  obtained by unrolling 485 time-steps from the first three time-steps of each of the 75 test trajectories. First row shows the vanilla neural network model, and the second row adds an interpolation projection layer to respect physical constraints imposed by the strings.}
    \label{tab:test_dyn}
\end{table}
\begin{figure*}
\centering
\begin{subfigure}[b]{.24\textwidth}
\includegraphics[width=\textwidth]{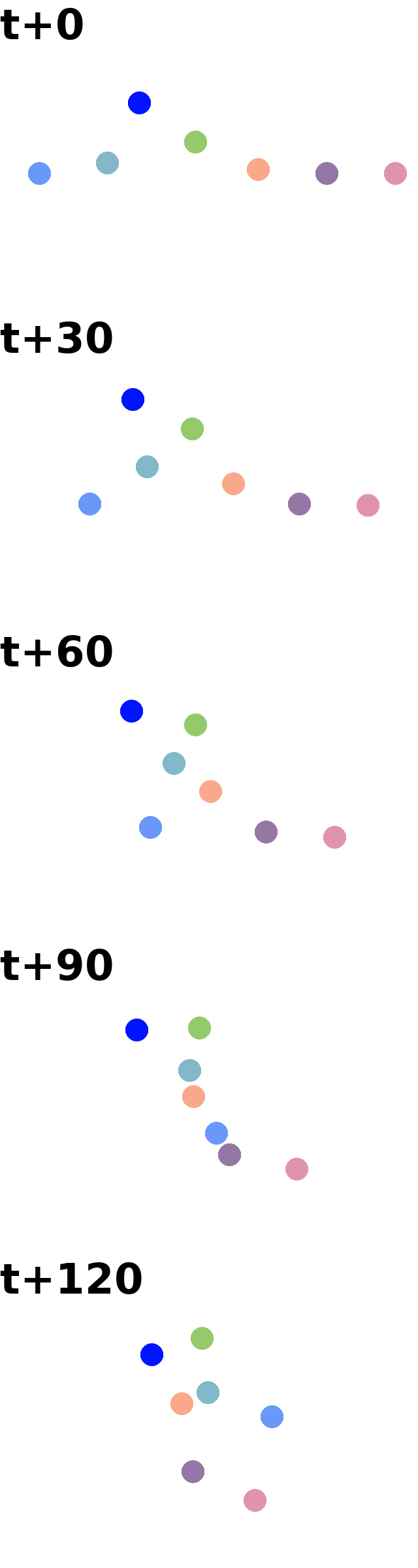}
\caption{Ground truth}
\end{subfigure}
\hfill
\begin{subfigure}[b]{.24\textwidth}
\includegraphics[width=\textwidth]{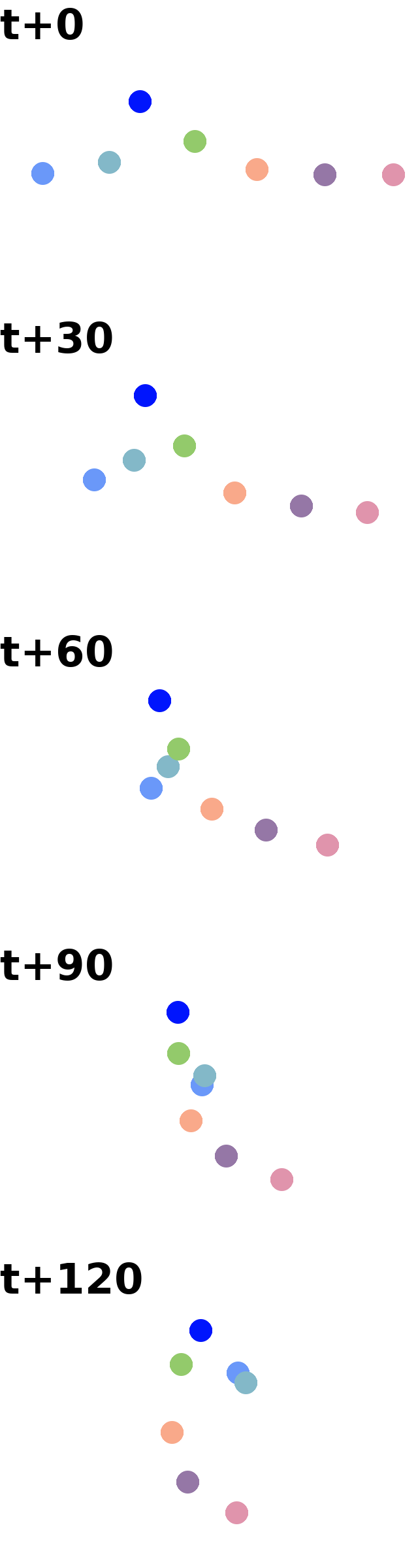}
\caption{RNN+Shape Cst.}
\end{subfigure}
\hfill
\begin{subfigure}[b]{.24\textwidth}
\includegraphics[width=\textwidth]{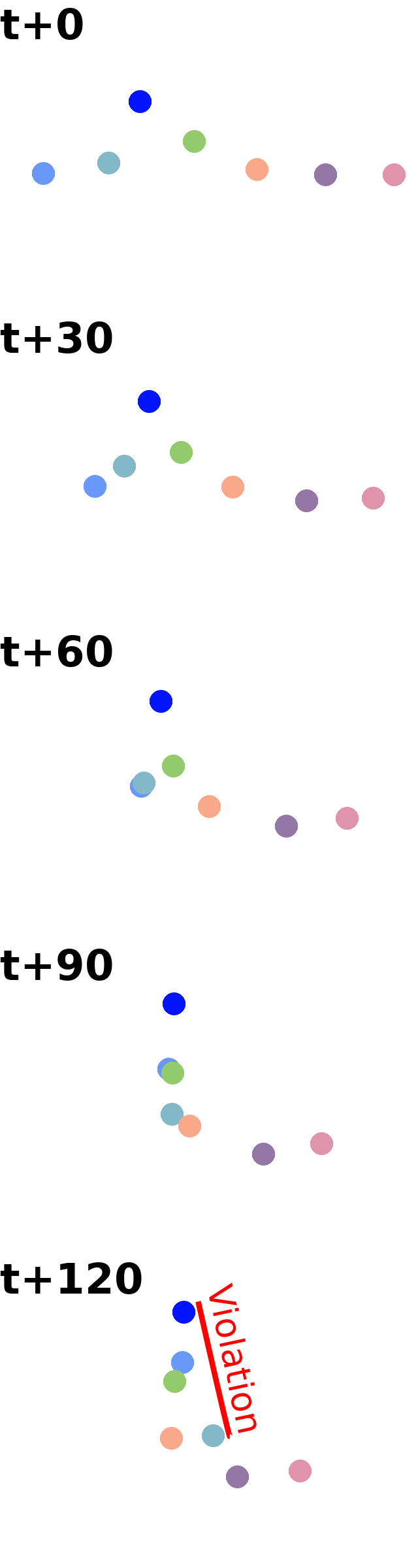}
\caption{RNN}
\end{subfigure}
\caption{Predicted trajectories vs ground truth. As errors compound, the RNN model without shape constraints exhibits large violations of the physical structure of the chain, as highlighted in red. In contrast, the model with the projection layer maintains physical consistency with the original shape at all times.}
\label{fig:qual1}
\end{figure*}


In the previous experiment we have shown how the interpolation projection can be used to tackle constrained optimization problems in the context of RL. In this experiment, we provide an example of an inductive bias in the form of a convex constraint on the outputs of a neural network, and we show how the interpolation projection can be used to comply with these constraints. The task consists in predicting the position, for several steps in the future, of 7 circular rigid bodies connected in 3 different configurations with respectively 6, 9 and 12 strings of the same length as shown in Fig.~\ref{fig:shapes}.

The considered inductive bias constrains the distance between predicted positions of connected rigid bodies to be at most the length of the string. To comply with the constraint, we add after the prediction of the neural network $y_t$, an interpolation projection that returns $g(y_t)$, such that the constraints imposed by the strings are respected. To compute $g$, we define $h$ as the maximum distance between linked bodies, which is convex, and use as `$x_0$'---the anchor point of the interpolation projection---an imaginary configuration that places all rigid bodies in the average of their position according to $y_{t-1}$. This point has thus zero distance between all circular bodies and strictly satisfies the constraints. Given $h$ and `$x_0$', the interpolation projection $g$ follows as in Sec.~\ref{sec:alg}. 

To predict the next position we use a neural network with 4 hidden layers having 256 nodes each. The network takes as input the last three positions of all 7 circular bodies and outputs the change to the current set of positions. We train this neural network as a recursive neural network (RNN), using backpropagation through time, as the predicted position in the next time-step is fed back to its input. In addition to the base RNN model, we evaluate the same RNN with the inductive bias in the form of convex constraints as described above. Ground truth trajectories are generated by letting the object fall from a distance of 400 units of measure (u.m.), after applying an initial force generated by selecting a node uniformly at random then applying a force with constant norm sampled uniformly at random on an upper half circle. The diameter of the circular rigid body is 1~u.m. Box2d~\citep{box2d} is used to simulate 200 of such trajectories, 50 of which are used for training, 75 for validation and 75 for test. Each trajectory contains 485 time-steps and the train set alone contains circa 24K time-steps. We train both the RNN and RNN with convex constraints for a fixed time of 3 days on a single core of an AMD 3900x. 

The generalization results in Tab.~\ref{tab:test_dyn} show that both models can synthesize relatively close trajectories to the original ones for an extended period of time (485 time-steps at 60Hz) from only the first three time-steps of the test trajectories. The results also show that the additional interpolation projection layer, enforcing compliance with the physical constraints imposed by the  strings, reduces the prediction error for the two shapes with the most strings; while for the simpler chain shape, the vanilla model performs better. The worse performance in this setup might be the result of the additional non-smoothness introduced by the interpolation projection. Yet, even when it under-performs quantitatively with the chain shape, the trajectories generated by the projection augmented model can look qualitatively better since the vanilla model sometimes exhibits large violations of the constraints as shown in Fig.~\ref{fig:qual1}. In conclusion, introducing an inductive bias through additional constraints and using the interpolation projection to comply with the constraints showed promising results both quantitatively and qualitatively, with little computational overhead---the training procedure becoming only about 1.2 times slower. In comparison, we were unable to run the baseline with the optimal projection layer that solves a convex problem for every forward pass. Compared to the RL setting, the combined effect of a larger dataset~(more than 10x) and the increased number of convex problems to solve per gradient update~(up to 240x du to the back-propagation through time) would require several months for the training procedure to complete on the same AMD 3900x processor. 

\section{Conclusion}
We introduced in this paper an interpolation-based projection onto a convex set that can be readily computed for any convex domain defining function. We then derived a descent algorithm based on the composition of the objective and the projection and showed that this surprisingly yields a convergent algorithm when the objective is linear, despite the `sub-optimality' of the projection. From a practical point of view, we have shown that this projection when added as a layer to computational models, allows to tackle constrained optimization in reinforcement learning or adds an inductive bias to predictive models. Because the projection is general and computationally frugal, we think this work can find many other applications in machine learning where intermediary nodes of a computational graph are constrained to be in a convex set.

\bibliographystyle{unsrtnat}      
\bibliography{references}

\newpage
\section*{Appendix A. Convex optimization numerical illustration}
We describe in more details the experimental setting of the convex optimization comparisons. We consider five problem classes comprising linear programs, semi-definite programs, second order cone programs, problems with a bounded $\ell_2$ norm and problems with an exponential form constraint. The form of the domain defining function $h$ for each of these problems is trivial except for the semi-definite cone, where we used $h(A) = -\lambda_{\min}$, the negative of the smallest eigenvalue of the symmetric real valued matrix $A$. The sub-gradient of $h$ w.r.t. $A$ is given in this case by $-vv^T$, where $v$ is the eigenvector associated with $\lambda_{\min}$. We now detail each problem class and its random instance generation.

\textbf{Linear program} (\texttt{Lin}).  The problem is 
\begin{equation*}
\begin{aligned}
\min_x \quad & c^T x,\\
\text{s.t.} \quad & a_i^T x \leq 0, \ i \in \{1 \ldots M\}.
\end{aligned}
\end{equation*}

We generate instances such that the optimum is at $(0, \ldots, 0)^T$ and the constraints are active at the optimum. The objective is generated by sampling a $c$ uniformly at random on the hyper-sphere. Following the idea in~\cite{coco,coco-code}, we define the constraints of such problems by setting the gradient of the first constraint to $a_1 = -c$ to ensure the Karush-Kuhn-Tucker optimality conditions~\cite{kuhn1951,nocedal2006} hold at $(0, \ldots, 0)^T$. At this point, the point $x=c$ is feasible and we generate the remaining $M - 1$ constraints randomly while making sure that $x$ remains feasible. Specifically, each $a_i$, for $i \in \{2 \ldots M\}$, is sampled on the hypersphere uniformly at random and redefined as $a_i = -a_i$ if $a_i^Tx > 0$.

\textbf{Semi-definite program} (\texttt{SDP}). The dual of the problem is given by
\begin{equation*}
\begin{aligned}
\min_x \quad & c^T x,\\
\text{s.t.} \quad & \sum_i x_i A_i \succeq C.
\end{aligned}
\end{equation*}
The constraint implies that $\sum_i x_i A_i - C$ is a positive semi-definite matrix. We generate the problem data following the code of \cite{malick09} to obtain problems where strong duality holds. There is one difference in the generation of the matrices $A_i$, that are made sparse in the original code, while we use $A_i = \frac{1}{2}(B_i + B_i^T)$ with entries of $B_i$ sampled from the Normal distribution.

\textbf{Second order cone program} (\texttt{SOC}).  The problem is 
\begin{equation*}
\begin{aligned}
\min_x \quad & c^T x,\\
\text{s.t.} \quad & \norm{A_ix + b_i}_2 \leq z_i^t x + d_i, \ i \in \{1 \ldots M\}.
\end{aligned}
\end{equation*}

The objective is generated by sampling a $c$ uniformly at random on the hyper-sphere. Then an $x_0$ is generated following the same procedure. All other problem data are then sampled from the normal distribution except $d_i$ that is computed such that $h(x_0) = 0$, i.e. $d_i = \norm{A_ix + b_i}_2 - z_i^t x$.

\textbf{Norm constraint} (\texttt{Norm}). The problem is
\begin{equation*}
\begin{aligned}
\min_x \quad & c^T x,\\
\text{s.t.} \quad & \norm{x}_2 \leq 1.
\end{aligned}
\end{equation*}
A random instance of the problem is generated by sampling a vector $c$ uniformly at random on the hyper-sphere such that the optimum $x^*$ is $-c$ with value $f(x^*)=-1$.

\textbf{Exponential constraint} (\texttt{Exp}) The problem is
\begin{equation*}
\begin{aligned}
\min_x \quad & c^T x,\\
\text{s.t.} \quad & \frac{1}{2}\norm{x-b}_2^2+\sum_{i=0}^{d-1}\exp(x_i-b_i) \leq d,
\end{aligned}
\end{equation*}
where $b$ is a vector that has on each entry $W(1)$, the Lambert W function evaluated at 1. It is designed such that the minimum of the constraint is attained at $(0, \ldots, 0)^T$, facilitating the generation of feasible points. $c$ is generated by sampling uniformly at random on the hyper-sphere.

\begin{figure*}
    \centering
    \def\figBenefitsWidth{.19}
        \includegraphics[width=\figBenefitsWidth\textwidth]{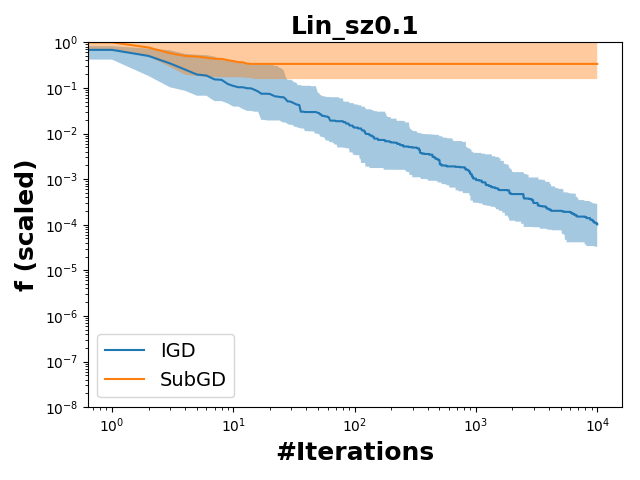}
        \includegraphics[width=\figBenefitsWidth\textwidth]{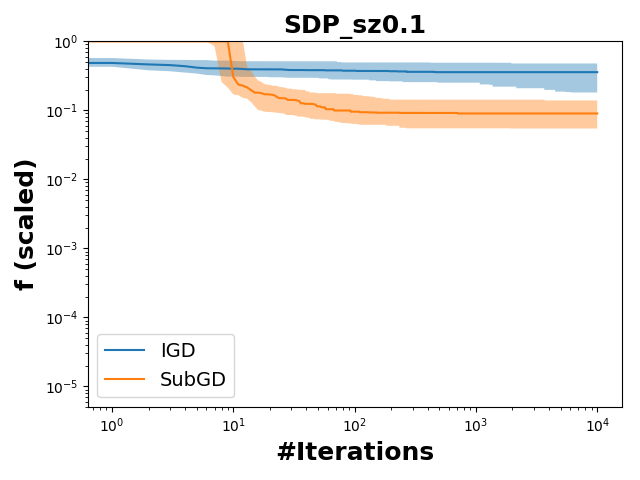}
        \includegraphics[width=\figBenefitsWidth\textwidth]{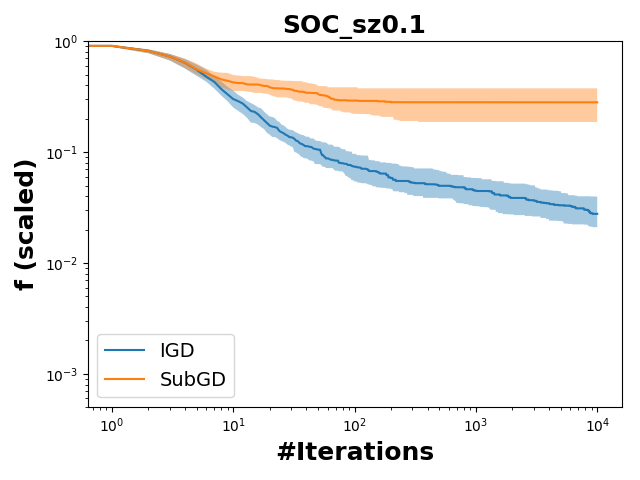}
        \includegraphics[width=\figBenefitsWidth\textwidth]{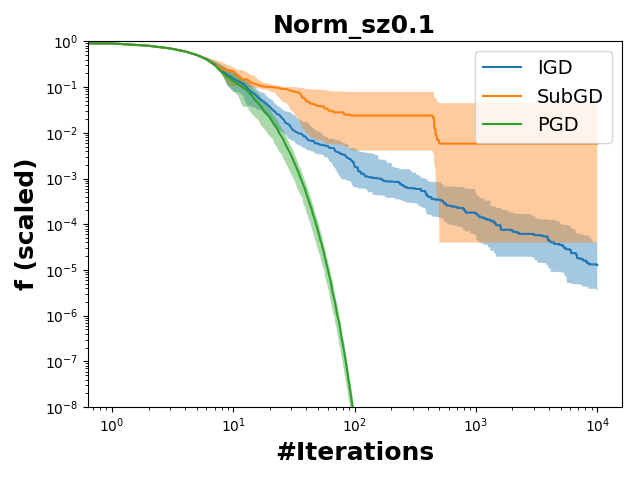}
        \includegraphics[width=\figBenefitsWidth\textwidth]{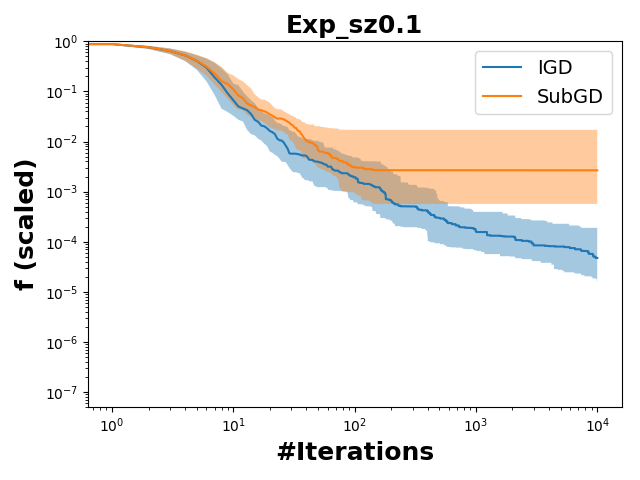}

        \includegraphics[width=\figBenefitsWidth\textwidth]{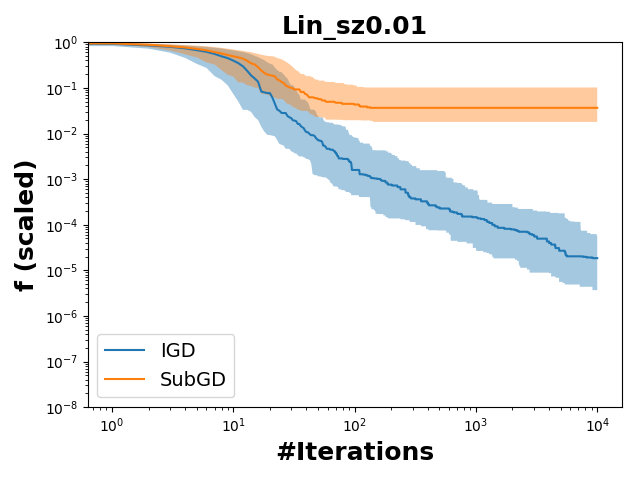}
        \includegraphics[width=\figBenefitsWidth\textwidth]{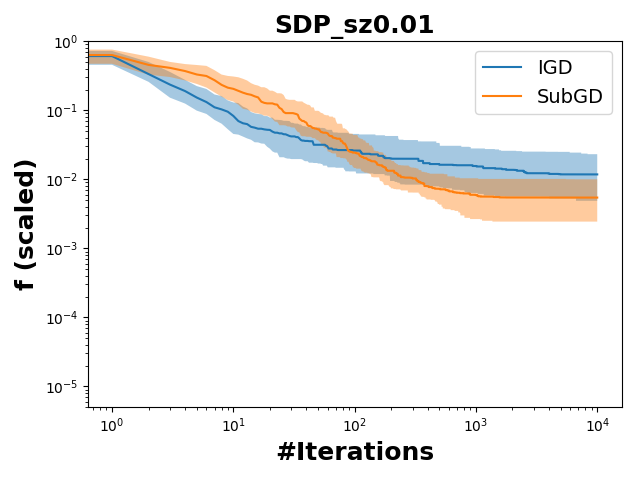}
        \includegraphics[width=\figBenefitsWidth\textwidth]{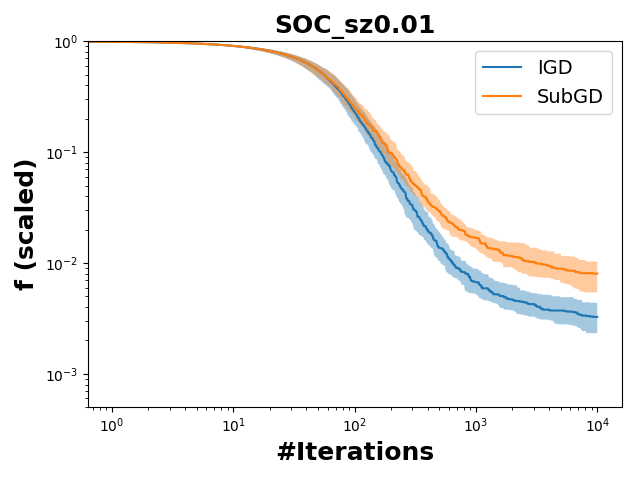}
        \includegraphics[width=\figBenefitsWidth\textwidth]{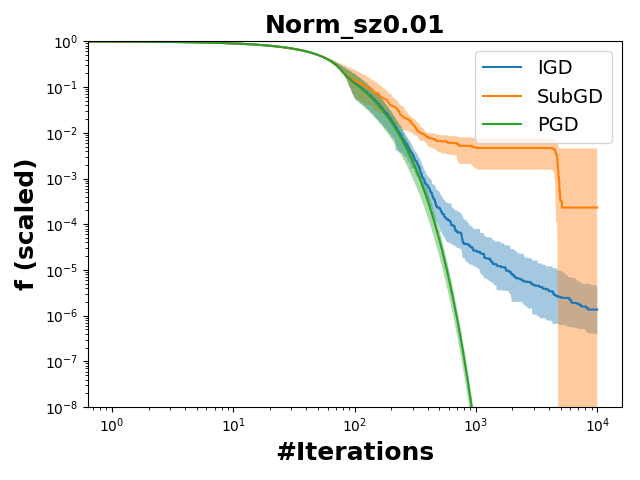}
        \includegraphics[width=\figBenefitsWidth\textwidth]{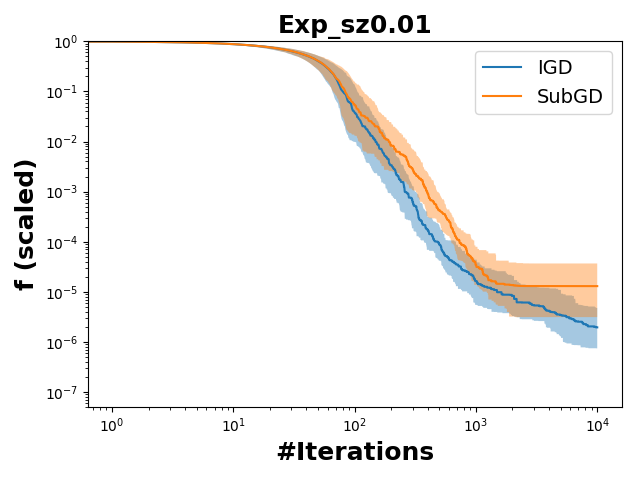}
              
        \includegraphics[width=\figBenefitsWidth\textwidth]{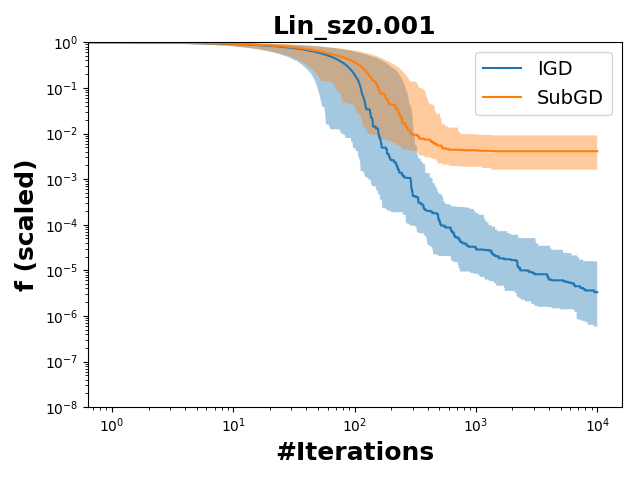}
        \includegraphics[width=\figBenefitsWidth\textwidth]{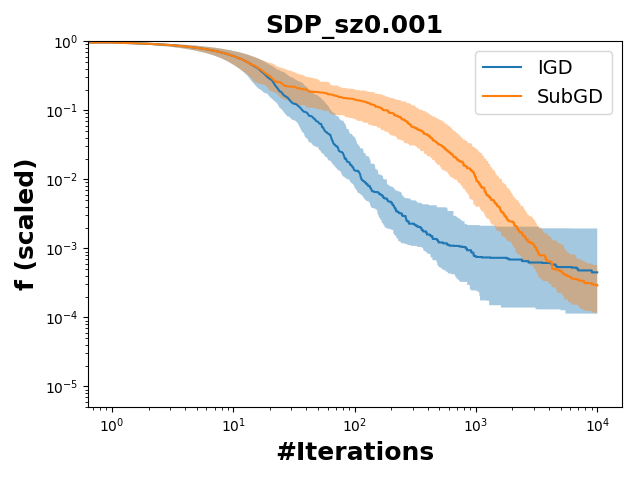}
        \includegraphics[width=\figBenefitsWidth\textwidth]{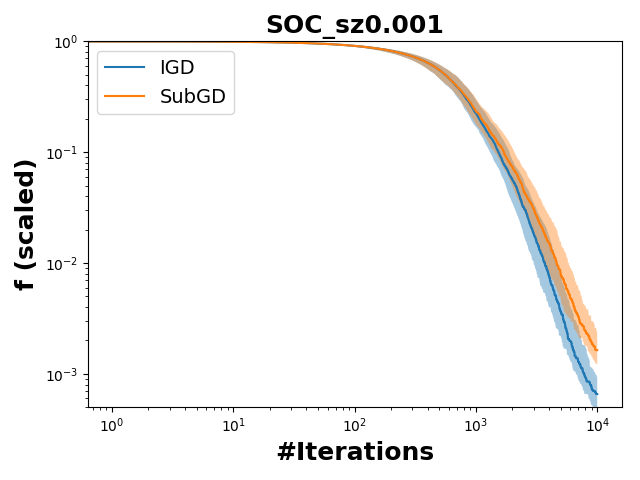}
        \includegraphics[width=\figBenefitsWidth\textwidth]{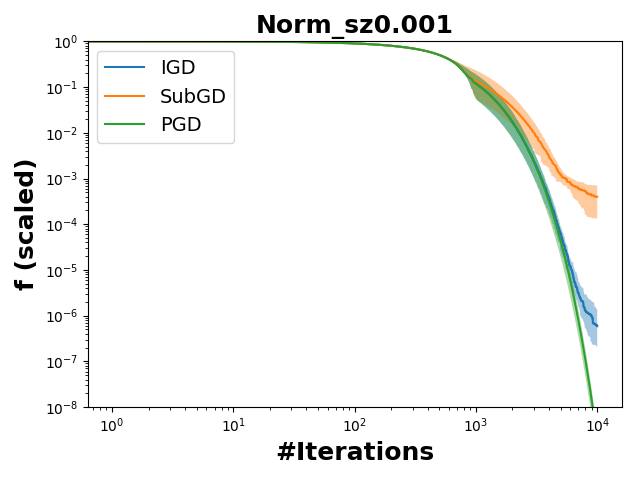}
        \includegraphics[width=\figBenefitsWidth\textwidth]{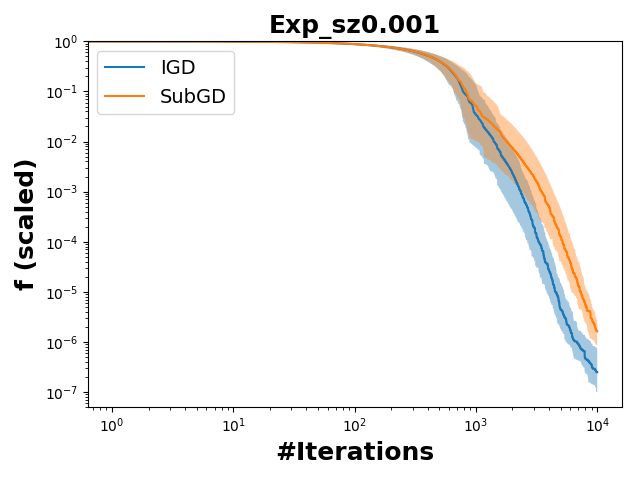}
        
         \includegraphics[width=\figBenefitsWidth\textwidth]{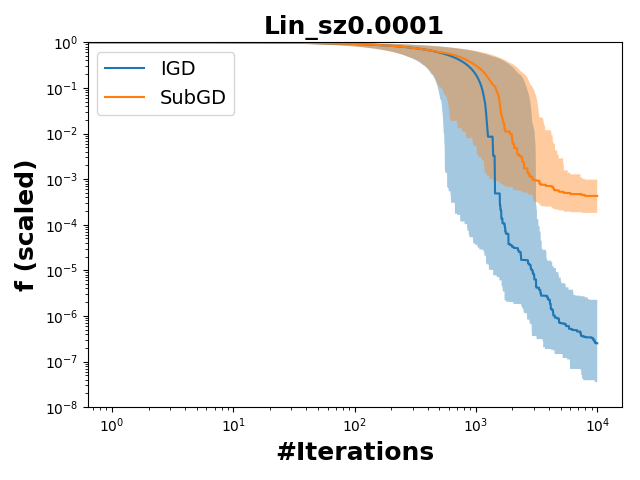}
        \includegraphics[width=\figBenefitsWidth\textwidth]{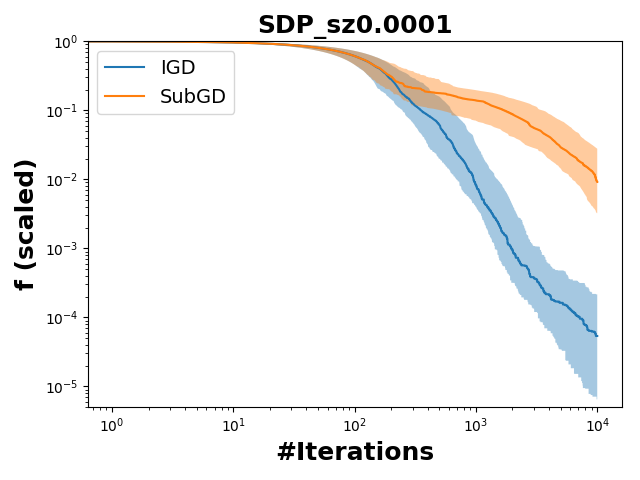}
        \includegraphics[width=\figBenefitsWidth\textwidth]{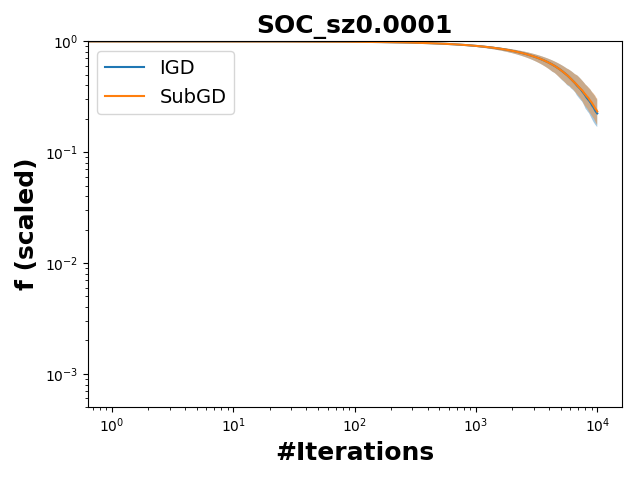}
        \includegraphics[width=\figBenefitsWidth\textwidth]{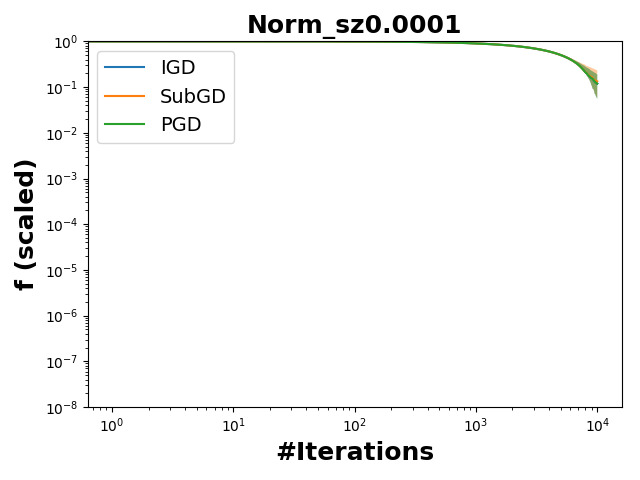}
        \includegraphics[width=\figBenefitsWidth\textwidth]{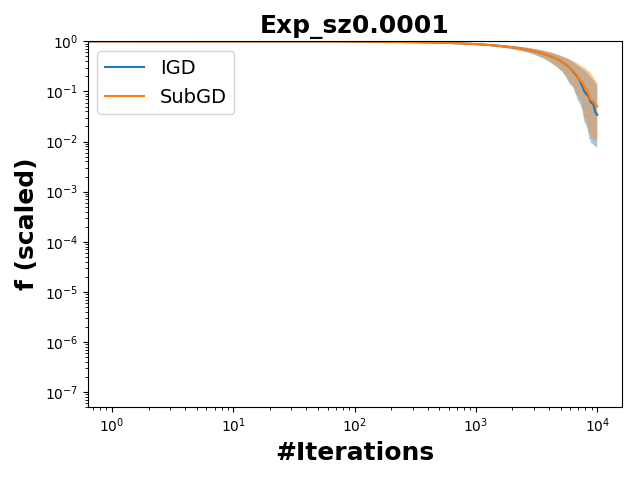}
    \caption{Comparison of first order descent algorithms with different step-sizes on linear programs (leftmost column), semidefinite programs, second order cone programs, programs with bounded norm or exponential shaped constraint (rightmost column). Step-size $\beta$ ranges from 0.1 on the first row to $10^{-4}$ on the forth row. All plots averaged over 100 runs.} 
    \label{fig:res}
\end{figure*}

\textbf{Obtaining $x_0$ and $f(x^*)$.} For \texttt{Lin}, \texttt{Norm} and \texttt{Exp}, $x_0$ is generated by uniformly sampling at random in the unit ball, and resampling if the point is not feasible. For \texttt{SDP} we use $x_0$ as in the code of \cite{malick09}. For \texttt{SOC}, our algorithm cannot use the $x_0$ described in the problem definition, since $h(x_0) = 0$. To obtain a valid $x_0$ for our algorithm, starting from the aforementioned $x_0$, we perform 100 optimization steps with Adam \cite{KingmaB14} and a step-size of $10^{-2}$ on the maximum over the constraints, and use the newly obtained point as the $x_0$ for all algorithms. For \texttt{Lin} and \texttt{Norm}, $f(x^*)$ is known whereas we estimate it for the remaining problems using CVXPY~\cite{diamond2016cvxpy} with the highest precision available.

\textbf{Performance metrics.}
For every optimization problem we randomly generate an instance and run all optimizers for 10000 iterations. We repeat this procedure 100 times for every problem. For each run, and at each iteration $k$, we compute $\min_{t\in\{1..k\}} f(g(x_t))$ where $g$ is the norm minimizing projection for PGD or the interpolation projection for our algorithm. For subgradient descent we use instead $\min_{t\in\{i \in \{1..k\} \text{s.t.} h(x_i)\leq 0\}} f(x_t)$, i.e. we pick the best point so far that is in $\cal C$. We consider the $\min$ instead of $f(\frac{1}{k}\sum_{t=0}^kg(x_t))$ as an evaluation metric for our algorithm in order to allow for comparisons with the subgradient descent method in which the average point so far, is not necessarily in $\cal C$. Note that the theoretical guarantees given by Thm.~\ref{thm} are exactly the same for this $\min$ criterion since $\min_{t\in\{1..k\}} f(g(x_t)) \leq \frac{1}{k}\sum_{t=0}^kf(g(x_t))$ can be used in a similar way in the proof in lieu of the average point. In order to allow for meaningful averaging between the several randomly generated instances, we normalize the performance between 0 and 1 for each run by subtracting $f(x^*)$ and dividing by $f(x_0)-f(x^*)$. Instances of \texttt{Lin}, \texttt{SDP}, \texttt{SOC} and \texttt{Norm} and \texttt{Exp} are of dimensionality 10, 10, 20, 100 and 2 respectively. For each problem, we evaluated all algorithms with step-sizes $\beta$ of $10^{-4}$, $10^{-3}$, $10^{-2}$ and $10^{-1}$. Random instances across different step-sizes are identical and results are therefore directly comparable. Finally, the performance plots in Fig.~\ref{fig:res} are obtained by plotting the median and the upper and lower quantiles. 

\textbf{Results.} On the plots of Fig.~\ref{fig:res}, one can notice on all problems that the performance of all algorithms perfectly overlaps in initial iterations. That is due to the fact that all compared algorithms are similar up to the point where an iterate first exits the feasible set $\cal C$. The plots also show that in 17 out of the 20 problem and step-size combination, IGD outperforms SubGD, sometimes with several order of magnitude. On semi-definite programs, SubGD performs better with larger step-sizes, although best results are still obtained overall by IGD with the smallest step-size. On the \texttt{Norm} problem where PGD is applicable and with $\beta=0.001$, we observe that both PGD and IGD perform very similarly despite the simplicity and the linear nature of the projection used by our algorithm, and both algorithms perform better than the more naive SubGD baseline. On these problems, our algorithm is able to match PGD up until a precision ranging from $10^{-2}$ to $10^{-5}$ for different step-sizes, before tracking behind. In contrast SubGD is distanced at a significantly lower precision. All combined, these results both demonstrate a certain robustness to the choice of step-size and a practical interest in the mixing of gradients obtained by differentiating through $f\circ g$. Thanks to the generality of the projection and the simplicity of performing unconstrained gradient descent on $f\circ g$, we expect the interpolation projection to find many usages in machine learning.

\end{document}